\documentclass[letterpaper]{article} % DO NOT CHANGE THIS
\usepackage{aaai24}  % DO NOT CHANGE THIS
\usepackage{times}  % DO NOT CHANGE THIS
\usepackage{helvet}  % DO NOT CHANGE THIS
\usepackage{courier}  % DO NOT CHANGE THIS
\usepackage[hyphens]{url}  % DO NOT CHANGE THIS
\usepackage{graphicx} % DO NOT CHANGE THIS
\usepackage{natbib}  % DO NOT CHANGE THIS AND DO NOT ADD ANY OPTIONS TO IT
\usepackage{caption} % DO NOT CHANGE THIS AND DO NOT ADD ANY OPTIONS TO IT
\usepackage{algorithm}
\usepackage{algorithmic}
\usepackage{booktabs}
\usepackage{adjustbox}
\usepackage{xcolor}
\usepackage{bm}
\usepackage{multirow}
\usepackage{threeparttable}
\usepackage{eso-pic}
\usepackage{array}
\usepackage{bbding}
\usepackage{amsmath}
\usepackage{newfloat}
\usepackage{listings}
\DeclareCaptionStyle{ruled}{labelfont=normalfont,labelsep=colon,strut=off} % DO NOT CHANGE THIS
\floatstyle{ruled}
\newfloat{listing}{tb}{lst}{}
\floatname{listing}{Listing}
\nocopyright
\title{CityPulse: Fine-Grained Assessment of Urban Change with\\ Street View Time Series}
\author{
    Tianyuan Huang\textsuperscript{\rm 1}\equalcontrib,
    Zejia Wu\textsuperscript{\rm 2}\equalcontrib,
    Jiajun Wu\textsuperscript{\rm 1},
    Jackelyn Hwang\textsuperscript{\rm 1},
    Ram Rajagopal\textsuperscript{\rm 1}
}
\affiliations {
    \textsuperscript{\rm 1}Stanford University
    \textsuperscript{\rm 2}University of California San Diego\\
    \{tianyuah, jihwang, ramr\}@stanford.edu, 
    zew024@ucsd.edu, 
    jiajunwu@cs.stanford.edu
}
\usepackage{bibentry}
\newtheorem{definition}{Definition}
\begin{document}

\maketitle

\begin{abstract}
Urban transformations have profound societal impact on both individuals and communities at large. Accurately assessing these shifts is essential for understanding their underlying causes and ensuring sustainable urban planning. Traditional measurements often encounter constraints in spatial and temporal granularity, failing to capture real-time physical changes. While street view imagery, capturing the heartbeat of urban spaces from a pedestrian point of view, can add as a high-definition, up-to-date, and on-the-ground visual proxy of urban change. We curate the largest street view time series dataset to date, and propose an end-to-end change detection model to effectively capture physical alterations in the built environment at scale. We demonstrate the effectiveness of our proposed method by benchmark comparisons with previous literature and implementing it at the city-wide level. Our approach has the potential to supplement existing dataset and serve as a fine-grained and accurate assessment of urban change.

\end{abstract}

\section{Introduction}
% \textbf{Significance of the problem: The social impact problem considered by this paper is significant and has not been adequately addressed by the AI community}\\
% Urbanization has led to profound changes in the physical, social, and economic landscapes of cities worldwide. 
Our cities are evolving, and understanding how cities change at a granular level has far-reaching societal impact --- from facilitating better urban planning and infrastructure assessment to enabling more sustainable social and environmental interventions \cite{Daniel2015Goal1M,Seto2017SustainabilityIA}. 
Current measurements of urban change rely on datasets ranging from survey data such as American Community Survey (ACS), to government open data like construction permits, to remote sensing data such as satellite and aerial imagery. However, survey data often fall short of spatial and temporal granularity \cite{hwang14}, and top-down perspectives from the remote sensing data may not adequately represent the street-level changes that directly impact the daily lives of urban residents. And some construction permits data are not universally accessible.
Street view imagery, on the other hand, offers a high-resolution and frequently updated visual representation of urban environments from a ground-level perspective \cite{Huang2022DetectingNG}. By curating and analyzing the time series data of street view imagery, we can establish a more precise and direct proxy for how cities evolve over time.
\begin{figure}[ht!]
    \centering
    \includegraphics[width=1.0\linewidth]{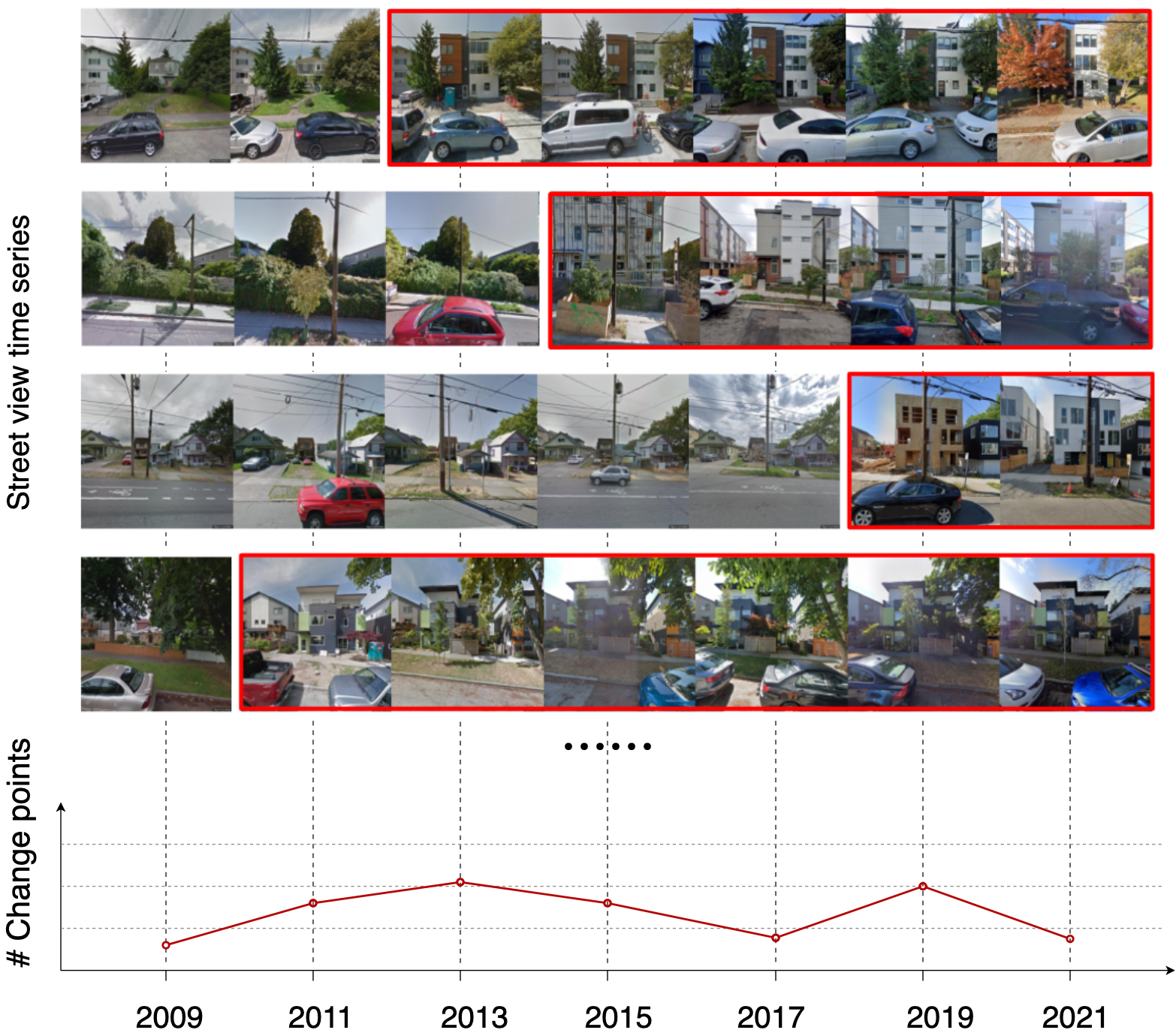}
    \caption{Detection of urban change points using street view time series. Red bounding boxes highlight transformations in the built environment at each location. By aggregating these detected change points within a neighborhood, we can evaluate the temporal dynamics of urban development.}
    \label{fig:main}
\end{figure}

% \begin{figure}[ht!]
%     \centering
%     \includegraphics[width=1.0\linewidth]{figure/main.png}
%     \caption{Detecting urban change points from street view time series. Red bounding boxes denote the change of built environment in sampled coordinates. By aggregating the total number of change points inside the neighborhood, we could assess how cities change over time.}
%     \label{fig:main}
% \end{figure}
Previous studies on street view change detection have primarily focused on comparing pairwise images from identical locations but at different times, utilizing pixel-level annotations \cite{Sakurada2015ChangeDF,sakurada2020weakly}, which is similar to change detection tasks using satellite imagery \cite{Leenstra2021SelfsupervisedPE,Shi2021ADS}. Recent works have also demonstrated the applicability of street view pairwise change detection by collecting large-scale historical street view datasets and apply them on a range of urban applications, such as mapping out physical improvements and declines in cities, as well as correlating with socio-economic attributes and neighborhood gentrification status \cite{Naik7571, Huang2022DetectingNG}. 

However, unlike satellite imagery, street view imagery can be more susceptible to noisy signals, such as varying camera angles and noisy background elements like shadows and lighting changes. Additionally, existing street view datasets for change detection tasks are often limited in both spatial and temporal scales due to the fact that pixel-level semantic annotations can be costly. Such constraints hamper the model's generalizability and scalability, making it challenging to directly apply them to downstream tasks and thus restricting their potential social impacts. 
% Additionally,  In, these works fall short in providing an open, transferable model and corresponding benchmark datasets. 

To address these challenges, we first introduce a comprehensive multi-city street view time series dataset with image-level semantic labels, and then propose an end-to-end framework to detect urban change with street view data. We demonstrate the effectiveness of our approach with a fine-grained assessment of urban change across Seattle, Washington. Specifically, our major contributions in this study are threefold: 
\begin{itemize}
    \item We collect and curate a Google Street View (GSV) time series dataset, covering more than $1000$ coordinates across $6$ different cities, which is the largest street view change detection dataset available up to date. Each street view time series is labeled with change or no change on the image level, and each series has an average length of $10$ images, covering a time interval of $16$ years (from $2007$ to $2023$). We further validate the benefits of the time series data over pairwise data in our experiment.
    \item We propose an end-to-end change detection pipeline that effectively learns feature representations with semantic contexts from street view time series data, which allows the model to not only extract object shape, color, and structural information of the built environment, but also mitigate noisy effects from lighting changes and angle misalignment, enhancing the overall robustness of the change detection.
    \item Our method enables scalable applications for urban scene change detection, providing a more accurate proxy for assessing neighborhood socio-economic status changes. We demonstrate the efficacy of our approach by evaluating its correlation with social-demographic data and comparing against construction permits through a case study in Seattle.
\end{itemize}
% Figure showing the assessment visualization \todo{Figure}.

\section{Related work}
% \textbf{Engagement with literature: Shows an excellent understanding of other literature on the problem, including that outside computer science.}
\subsection{Urban change assessment}
Measuring physical change in urban environments offers profound insights into urban policies and economics, illuminating housing value trends, shifts in urban areas' roles, and spatial segregation effects \cite{Temkin1996NeighborhoodCA,Hwang2020UnequalDG}. It also has significant values in a various downstream tasks, such as detecting neighborhood gentrification \cite{hwang_gentrification}, monitoring the disaster recovery \cite{Stevenson2010UsingBP}. Prior research primarily utilizes satellite imagery for large-scale urban change detection \cite{Pandey2015UrbanizationAA,Daudt2018UrbanCD}. However, satellite data are hindered by constraints in spatial and temporal resolution, and lack detail on fine-grained and street-level changes. Several researchers use building permits data as a fine-grained proxy for physical urban change \cite{Stevenson2010UsingBP,Strauss2013DoesHD}. While these data also have limitations in availability and spatial coverage and may not accurately represent actual changes due to potential delays, as indicated by our evaluations.

\subsection{Street view imagery}
Street view imagery have been used in a wide range of applications in urban studies, such as quantifying urban greenery \cite{Li2015AssessingSU}, indicating region functions \cite{Gong2019ClassifyingSS}, revealing economic and social-demographic patterns\cite{gebru2017using,wang2020urban2vec,tian2021}, predicting populace' well-beings \cite{Lee2021PredictingLI} and estimating building energy efficiency \cite{MAYER2023120542}. 
It demonstrates substantial value as a medium closely reflecting human perception of the city. 
Moreover, recent studies analyzed historical street view data in the temporal dimension to map physical improvement and decline in the built environment and uncover how cities changed over time \cite{Naik2014StreetscoreP,Naik7571,Huang2022DetectingNG}. However, existing methods rely on comparing pairs of historical street views for each location rather than a comprehensive time series of street view data, which is limited in tracking the complete range of transformations within urban environments.

\subsection{Change detection}
Change detection is commonly presented in the field of remote sensing, as the task to identify changes between the pixel-level features of two temporally separated images. Previous works have trained convolutional network, recurrent network and Siamese network for change detection task on satellite imagery \cite{Gong2016ChangeDI,Daudt2018FullyCS,Lyu2016LearningAT}. Recent works also explored self-supervised pretraining and unsupervised learning in change detection to rely less on labels and generate meaningful representations for other downstream tasks \cite{Cong2022SatMAEPT,Mall2022ChangeED}. Unlike satellite imagery, Street view change detection faces additional challenges such as noisy signals including shades and angle misalignments due to the less fixed and more variable nature of acquiring street-level visual data. Previous works introduced benchmarking datasets for scene change detection and adopted deconvolutional networks or temporal attention networks \cite{Alcantarilla2016StreetviewCD,Sakurada2015ChangeDF,Chen2021DRTANetDR}. However, current research not only lacks comprehensive benchmark datasets with expansive spatial and temporal coverage, but also falls short of the model generalizability, limiting their societal impact. To tackle this, we introduce the largest image-level change detection dataset to date, featuring a complete time series of street views for each sampled location, applied at the city scale.

\begin{figure*}[h!]
    \centering
    \includegraphics[width=0.98\linewidth]{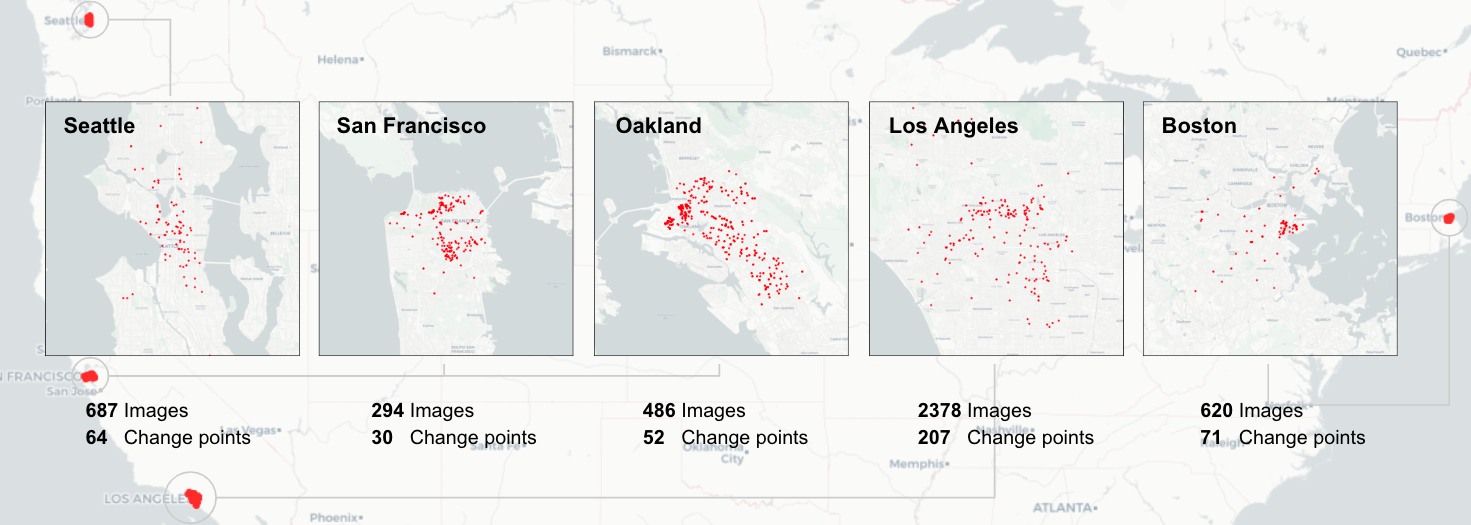}
    \caption{Geo-spatial distribution of our street view time series dataset across 5 different cities in the US. Locations are selected based on open-access building footprint data, and historical Google Street View imagery from these coordinates is comprehensively downloaded and labeled with urban change points.}
    \label{fig:data_geo}
\end{figure*}

% \begin{figure*}[h!]
%     \centering
%     \includegraphics[width=0.98\linewidth]{figure/data_geo.png}
%     \caption{Geo-spatial distribution of our street view time series dataset. We sample street view time series data across 5 different cities in the US. Each location is determined through open-access building footprint data, and we download all the historical Google Street View that recorded there.}
%     \label{fig:data_geo}
% \end{figure*}
% \subsection{Self-supervised Pretraining}
\section{Methods}
% \textbf{Novelty of approach:Introduces a new model, data gathering technique, algorithm, and/or data analysis technique.}
\subsection{Problem Statement}
\begin{definition}
\textbf{Street view time series.}
Each street view time series, comprises $n$ street view images depicting the consistent street-level scene $s^{(i)}=(s^{(i)}_{1}, s^{(i)}_{2}, ... , s^{(i)}_{n})$. These images are chronologically arranged such that $s^{(i)}_{k}$ corresponds to the timestamp $t^{(i)}_k$.
\end{definition}

\begin{definition}
\textbf{Urban change point.}
In the street view time series $s^{(i)}$, the image $s^{(i)}_{c}$ is identified as an urban change point if the built environment in $s^{(i)}_{c}$ exhibits deviations (e.g., building constructions) relative to preceding images.
\end{definition}

Our objective is to accurately and efficiently detect urban change points in street view time series. To achieve this, we begin by creating a large street view time series dataset, followed by proposing an end-to-end training and evaluation pipeline.

\subsection{Street view time series dataset}
To start, we sample the geospatial coordinate for each street-level scene $s^{(i)}$. Specifically, the coordinate of each building is determined by computing the centroid of its footprint polygon, using the Microsoft building footprint dataset.
After locating scene $s^{(i)}$, we gather all the available historical street view metadata and subsequently download the associated images using their panoid ID. For each scene $s^{(i)}$, we retrieve the nearest-photographed panorama. The image heading is subsequently determined, facing the building from the panorama's coordinate. All our street view images and meta data are sourced from the Google Static Street View API.
\begin{table}[h!]
  \centering
  \resizebox{0.9\linewidth}{!}{
    \begin{tabular}{lcclc}
    \toprule
    Dataset & \# images & \# pairs & Areas & Timeframe\\
    \midrule
    TSUNAMI & 200  & 100  & 2 cities  &  $<1$ year  \\
    PSCD & 1540  & 770 & 1 city & $<10$ years \\
    Ours & \textbf{4465}  & \textbf{25423} & 5 \textbf{cities} & \textbf{16 years} \\
    \bottomrule
    \end{tabular}}
  \caption{Scene change detection dataset comparison}
  \label{table:data_compare}
\end{table}

In total, we select $931$ locations and retrieve their corresponding street view time series, which consist of $10,878$ images. We then annotate each time series $s^{(i)}$ to identify the urban change points $s_c^{(i)}$. Among them, $371$ time series have been labeled with a total of $433$ urban change points, while the remaining ones exhibit no substantial urban change. Figure \ref{fig:data_geo} demonstrates the geo-spatial distribution of our sampled street views. 
As is shown in table \ref{table:data_compare}, our dataset not only consists of more image pairs compared with previous scene change detection datasets TSUNAMI \cite{Sakurada2015ChangeDF} and PSCD \cite{sakurada2020weakly}, but also covers a significantly broader spatial and temporal scope.

\subsubsection{Data partitioning.}
\begin{figure}
    \centering
    \includegraphics[width=1.01\linewidth]{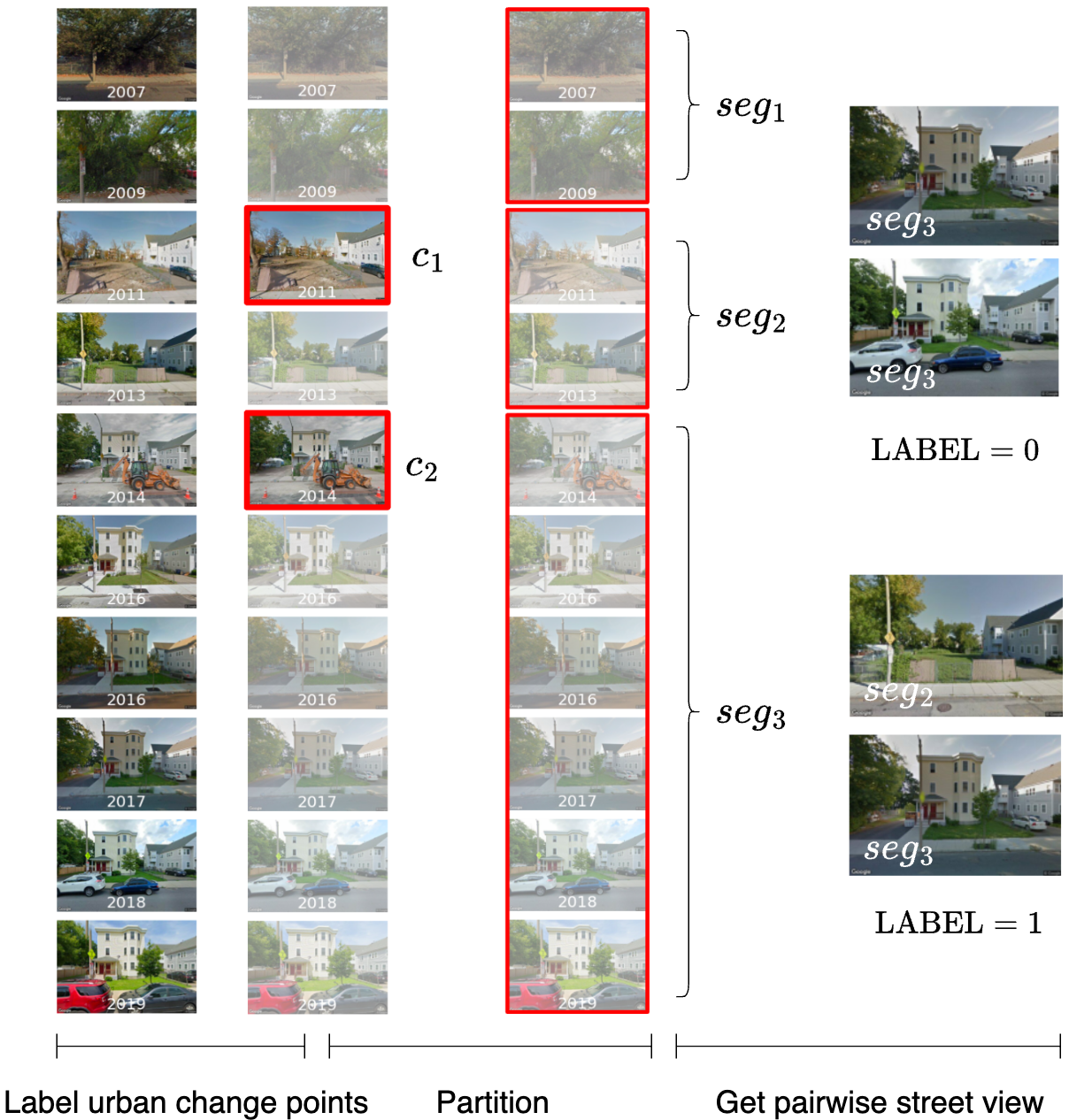}
    \caption{Partitioning of street view time series data. All possible pairwise combinations of street view samples are generated from each time series. Each pair's label is assigned based on its corresponding position with the urban change points.}
    \label{fig:data_method}
\end{figure}

To train our change detection model on street view time series dataset labeled with urban change points, we introduce a partitioning scheme for generating training and evaluation sets as is shown in Figure \ref{fig:data_method}. For every street view time series $s^{(i)}$, we segment the views from $s_1^{(i)}$ to $s_n^{(i)}$ based on their occurrence relative to $s_c^{(i)}$. More precisely, suppose the time series $s^{(i)}$ has $q$ urban change points denoted as $s_{c_1}^{(i)}$, $\ldots$, $s_{c_q}^{(i)}$, we allocate the street view $s_j^{(i)}$ ($1\leq j\leq n$) to segment $seg(s_j^{(i)})$ as follows:
\begin{equation}
    seg(s_j^{(i)})=
    \begin{cases}
    1 & \text{if}\ j<c_1\\
    k & \text{if}\ c_1\leq j < c_q\ \text{and}\ c_{k-1}\leq k<c_k\\
    q+1 & \text{if}\ j \geq c_q
    \end{cases}
\end{equation}

% \begin{figure}
%     \centering
%     \includegraphics[width=1.01\linewidth]{figure/data_method.png}
%     \caption{Street view time series data partitioning. We generate all the combinations of pairwise street view samples by partitioning each street view time series. Each pair's label is determined based on their corresponding position with the change point labels.}
%     \label{fig:data_method}
% \end{figure}
\begin{figure}[h!]
    \centering
    \includegraphics[width=1.05\linewidth]{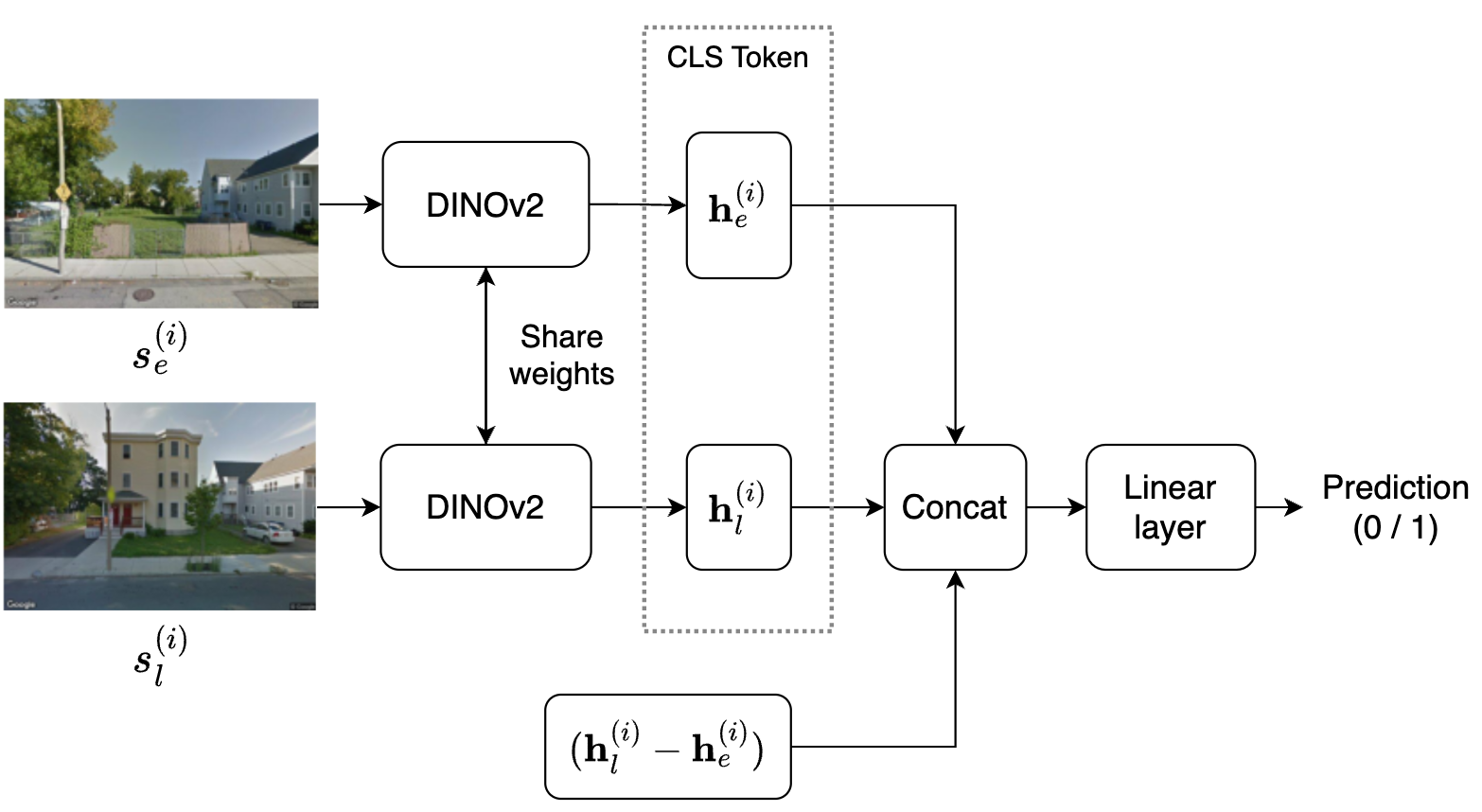}
    \caption{Overview of the change detection model architecture. Pairs of input images are processed using Siamese-based networks with DINOv2 as the backbone. The CLS tokens serve as the image representation, with a subsequent linear layer projecting them to a prediction score.}
    \label{fig:change_model}
\end{figure}

% \begin{figure}[h!]
%     \centering
%     \includegraphics[width=1.05\linewidth]{figure/change_model.png}
%     \caption{Overview of the change detection model architecture. Each pair of input images are passed through a Siamese-based networks with DINOv2 as backbone modules, we treat the CLS token as the representation of each image and add a linear layer to preject into a prediction score.}
%     \label{fig:change_model}
% \end{figure}
For each street view time series $s^{(i)}$, we then generate a set of pairwise street view pairs by considering all combinations from $s_1^{(i)}$ to $s_n^{(i)}$. Each pair of samples is sorted in chronological order based on their timestamps, resulting in pairs like $(s_1^{(i)}, s_2^{(i)})$. The total number of such combinations for the time series $s^{(i)}$ with $n$ street views is $\binom{n}{2}$. The labeling of these pairwise samples is determined by their associated segments as follows:
\begin{equation}
    \textsc{label}{(s_a^{(i)},s_b^{(i)})}=
    \begin{cases}
        1 & \text{if}\ seg(s_a^{(i)})\neq seg(s_b^{(i)})\\
        0 & \text{if}\ seg(s_a^{(i)}) = seg(s_b^{(i)})
    \end{cases}
\end{equation}
\subsection{Change detection model}
To classify each pairwise pair $(s_a^{(i)},s_b^{(i)})$, we adopt a Siamese network to include a twin DINOv2 \cite{Oquab2023DINOv2LR} backboned module to realize a non-linear embedding from the input domain and a final linear layer transforming the concatenation of both images' hidden vectors and their distance, represented by their element-wise difference, into a scalar predictor as follows:
\begin{equation} \label{eq: sia_concat}
\mathbf{h}^{(i)}_{L} = \left[(\mathbf{h}^{(i)}_{l, L})^\top, (\mathbf{h}^{(i)}_{e, L})^\top, (\mathbf{h}^{(i)}_{l, L}-\mathbf{h}^{(i)}_{e, L})^{\top}\right]^{\top}
\end{equation}
Figure \ref{fig:change_model} visualizes the model architecture. We adopt a cross-entropy loss function to train such a urban change classifier, and let $\textsc{label}{(s_a^{(i)},s_b^{(i)})}$ be the label for the street view pair $(s_a^{(i)},s_b^{(i)})$.

% in the following form \todo{finalize loss function}:
% \begin{multline} \label{eq: bce}
%     \mathcal{L_S}(t^{(i)}) = \mathbf{y}(t^{(i)})\log\mathbf{p}(t^{(i)})+\\ (1-\mathbf{y}(t^{(i)}))\log(1-\mathbf{p}(t^{(i)})).
% \end{multline}

\section{Experiments}
% \textbf{Justification of approach: Thoroughly and convincingly justifies the approach taken, explaining strengths and weaknesses as compared to other alternatives.}\\
% \textbf{Quality of evaluation: Evaluation was exemplary: data described the real world and was analyzed thoroughly.}
To evaluate the efficacy of our proposed method, we conduct experiments from three perspectives: 1) Backbone models --- we benchmark the performance of selected visual foundational models in the context of street view change detection tasks. 2) Street view time series data --- we employ experiments to substantiate the advantage of time series data as a natural form of data augmentation \cite{seco}, compared with results achieved through artificial data augmentation. 3) Self-supervised pre-training --- we explore 2 pre-train methodologies using a larger-scale unlabeled street view dataset in order to evaluate the performance of a domain-specific pre-trained models for our change detection task.
% Firstly, we evaluate the performance of diverse generic visual foundational models in the context of street view image change detection tasks. 
% Subsequently, we employ experiments to substantiate the advantage of time series imagery, a natural form of data augmentation within our collected dataset, which not only showcases superiority over pairwise datasets but also significantly outperforms results achieved through artificial data augmentation. Lastly, we introduce two pre-train methodologies, which are applied to a larger-scale unlabeled dataset of street-level images. This enables a comparative analysis between domain-specific pre-trained models and generic foundational models concerning their respective performances in the street view image change detection task.
\subsection{Training details}
% Our labeled dataset comprises a total of $371$ street view time series, comprised of $4,465$ images, sampled from $371$ distinct locations across $5$ different cities. By employing the data partitioning methodology described in Figure \ref{fig:data_method}, the dataset can be transformed into a collection of $25,423$ image pairs with binary labels. 
Street view images differ from satellite imagery and high-quality object images in that they often have a lower signal-to-noise ratio, primarily due to varying camera positions and environment conditions as shown earlier. As a result, evaluating on a small-scale test set could suffer from a significant variance. To ensure a robust assessment, we randomly select $50\%$ of the street view time series in our dataset as the test set. It includes $25$ locations in Seattle, $13$ locations in San Francisco, $21$ locations in Oakland, $97$ locations in Los Angeles, and $29$ locations in Boston, constituting a total of $12,221$ image pairs. The remaining data are allocated with $90\%$ as the training set for model fine-tuning and $10\%$ as the validation set.
% which to $185$ temporal sequences, as the test set. It includes $25$ locations in Seattle, $13$ locations in San Francisco, $21$ locations in Oakland, $97$ locations in Los Angeles, and $29$ locations in Boston, constituting a total of $12,221$ image pairs. The remaining data are allocated with $90\%$ as the training set for model fine-tuning and $10\%$ as the validation set.
During fine-tuning, we employ the Adam optimizer to train models with a learning rate set at $1 \times 10^{-5}$ and a batch size of $16$. The global norm of gradients is clipped to be $\leq 0.5$, and we use random weight averaging for optimization. Our training and evaluation are conducted on $4$ Nvidia Tesla T4 GPUs. For all the backbone models, we experimented with two common approaches: global fine-tuning and linear probing, i.e. training with the backbone network frozen.

\subsubsection{Backbone models.}
We initiate our evaluation by assessing the performance of 4 pre-trained generic visual models—ResNet101 \cite{he2016deep}, DINO \cite{dino}, CLIP \cite{radford2021learning}, and DINOv2 \cite{Oquab2023DINOv2LR}—as backbone networks. For ViT-based models such as DINO and DINOv2, we experiment using the CLS Token as the backbone output.
% For ViT-based models such as DINO and DINOv2, we experiment using either the CLS Token or the patch features as the backbone output, retaining the best results on the test set. When utilizing patch features as the output, we incorporate both a linear and a non-linear convolutional module ($\textit{Conv2d} \rightarrow \textit{BatchNorm} \rightarrow \textit{ReLU} \rightarrow \textit{Conv2d}$). Their outputs are then summed and dimensionally reduced to $100$ using a $1 \times 1$ convolutional kernel size.
% Specifically when using patch features as the output, we introduce a linear and a non-linear convolutional module, followed by the summation of their outputs, to reduce the dimensionality of the patch features to $100$, with a $1 \times 1$ convolutional kernel size. % The non-linear convolutional module comprises the sequence: $Conv2d \rightarrow BatchNorm \rightarrow ReLU \rightarrow Conv2d$.
% For the prediction layers after the backbone model, we explored both linear prediction layers and non-linear prediction heads composed of $\textit{Linear }\rightarrow \textit{BatchNorm} \rightarrow \textit{ReLU} \rightarrow \textit{Linear}$ sequences. 
% Table \ref{table:backbones} presents a comprehensive display of the best performances achieved by each respective backbone network on our test set.
\begin{table*}[t!]
  \centering
  \resizebox{0.8\linewidth}{!}{
  \begin{tabular}{l|cccc|cccc}
    \toprule
    Models & \multicolumn{4}{c|}{Linear Probing} & \multicolumn{4}{c}{Fine-Tuning} \\
    \cmidrule(r){2-5} \cmidrule(l){6-9}
    & Accuracy & Precision & Recall & F1-Score & Accuracy & Precision & Recall & F1-Score \\
    \midrule
    ResNet101 & 69.18 & 62.43 & \textbf{92.17} & 74.44 & 84.52 & 79.92 & \textbf{91.07} & 85.13 \\
    % DINO (ViT-B/16) & 86.86 & \textbf{92.96} & 78.99 & 85.41 & 87.76 & 90.68 & 83.43 & 86.90 \\
    DINO (ViT-B/16) & 82.33 & 82.24 & 81.25 & 81.74 & 86.24 & \textbf{93.30} & 77.28 & 84.54 \\
    CLIP & 86.01 & 86.95 & 83.85 & 85.37 & 87.06 & 91.52 & 80.91 & 85.89 \\
    DINOv2 (ViT-B/14) & \textbf{88.20} & \textbf{92.08} & 82.88 & \textbf{87.24} & \textbf{88.85} & 92.77 & 83.62 & \textbf{87.96} \\
    \bottomrule
  \end{tabular}}
  \caption{Performance of different backbone models using linear probing and fine-tuning.}
  \label{table:backbones}
\end{table*}

\begin{table*}[h!]
  \centering
  \resizebox{0.93\linewidth}{!}{
    \begin{tabular}{l|c|c|cccc}
    \toprule
    Data & Data Augmentation & \# pairs & Accuracy & Precision & Recall & F1-Score \\
    \midrule
    Pairwise data & None & 336 & 85.99 & 85.28 & \textbf{86.07} & 85.67 \\
    Pairwise data & HorizontalFlip + ColorJitter + GrayScale + GaussianBlur& 11922 & 85.63 & 88.60 & 80.89 & 84.57 \\
    Time series data & None& 11922 & \textbf{88.85} & \textbf{92.77} & 83.63 & \textbf{87.96} \\
    \bottomrule
    \end{tabular}}
  \caption{Street view time series vs. pairwise data. }
  \label{table:timeseries}
\end{table*}

\subsubsection{Time series data.}
% with DINOv2 as the backbone network, we utilize the architecture in Figure \ref{fig:change_model} to fine-tune the model. 
To validate the advantages of our proposed street view time series dataset, we constructed a pair-based dataset similar to TSUNAMI \cite{Sakurada2015ChangeDF} and PSCD \cite{sakurada2020weakly} dataset. Specifically, for each street view time series, we randomly sample $2$ images as a pair. We conduct model fine-tuning on this pair-based dataset and evaluated its performance on the test set described earlier.
% In sequences where changes occurred (positive sequences), we ensured the formation of one positive sample pair and one negative sample pair. This resulted in a pairwise dataset comprising $336$ sets of image pairs. 
To align the pair-based dataset with the size of our time series dataset, we randomly apply a combination of standard image augmentation techniques, including horizontal flip, color jitter, grayscale, and Gaussian blur. It seeks to validate our hypothesis that time series images, serving as natural augmentation, are more effective than artificial augmentations to supervise change detection model amidst noisy signals, thus bolstering its robustness.
% Assuming that we randomly sample $n$ pairs of images from time series $s^{(i)}$ (where $n\ mod\ 2 = 0$), the quantity for augmentation $a_{s^{(i)}}$ in $s^{(i)}$ can be determined using:
% \begin{equation}
% \begin{aligned}
%     a_{s^{(i)}} = \frac{n}{2} &\times [\mathrm{Ceil}(\frac{2 \times \text{\# positive image pairs}}{n})\\
%     &+\mathrm{Ceil}(\frac{2 \times \text{\# negtive image pairs}}{n})-2]\\
% \end{aligned}
% \end{equation}
% if $s^{(i)}$ is a positive sequence, or:
% \begin{equation}
% \begin{aligned}
%     &a_{s^{(i)}} = n \times [\mathrm{Ceil}(\frac{\text{\# image pairs}}{n})-1]
% \end{aligned}
% \end{equation}
% if $s^{(i)}$ is a negative sequence.

% We delved into different combinations of random data augmentation techniques and progressively increased the sampling quantity n from $2$ to $10$. This exploration aims to verify our hypothesis that, in contrast to artificial data augmentation, time series images as a form of natural augmentation can better enable the model to adapt to environmental noise, thereby enhancing its robustness. The results are presented in Table \ref{table:timeseries}.

%\subsection{Baselines}
%\begin{itemize}
%    \item[1] DINO (**freeze Resnet**, **freeze CLIP**, **DINO\_v2**) (DINO+?) (heading+batch norm) - zero-shot
%    \item[2] Time Series (vs random augmentation)
%    \item[3] Pretrain (BYOL, segmentation+BYOL, **SatMAE**) (**$10\%$ samples**) 
%\end{itemize}

\subsubsection{Self-supervised pre-training.}
% \todo{Model structure digram}\\
% With the advancement of Natural Language Processing (NLP), an increasing body of research is dedicated to the training of foundational models in Computer Vision. These models possess the capacity to generate generic visual features and enhance performance across downstream tasks. The majority of these foundational models are established through large-scale self-supervised training, 
Recent studies on self-supervised pre-training highlight its efficacy to extract image features when labels are limited and enhance performance in downstream tasks. Specifically, 2 primary branches of self-supervised pre-training are pursued: intra-image self-supervised training \cite{He2022masked,satmae2022} and discriminative self-supervised learning \cite{grill2020bootstrap}. Correspondingly, we adapt 2 pre-training procedures on street view data and benchmark their performance on the change detection task --- StreetMAE and StreetBYOL. StreetMAE uses masked autoencoders \cite{He2022masked} to reconstruct randomly masked patches in street view imagery. It also incorporates temporal encoding to represent each street view time series as a contextual sequence. StreetBYOL, on the other hand, is a self-distillation approach building upon the online and target networks \cite{grill2020bootstrap}. While retaining pivotal components such as the prediction head and the stop gradient mechanism, we try add an unsupervised segmentation head \cite{hamilton2021unsupervised} to identify building pixels and feed them alongside the original images into the networks in experiment seg+StreetBYOL. We adopt the ViT-B/16 architecture as the backbone network and initialize it with the parameters from DINO pre-trained model. Pre-training is conducted on an unlabeled dataset comprising 150,000 street view images randomly sampled in our studied areas. To mitigate noise interference, we apply a filtering process to remove images where the proportion of building pixels was less than $2\%$. After the pre-training phase, we plug it into the Siamese network and fine-tune the model on our labeled training set. 
\begin{table}[h!]
  \centering
  \resizebox{1.0\linewidth}{!}{
    \begin{tabular}{lcccc}
    \toprule
    Pre-training & Accuracy & Precision & Recall & F1-Score \\
    \midrule
    DINOv2 (ViT-B/14) & \textbf{88.85}  &  \textbf{92.77} & \textbf{83.62} & \textbf{87.96} \\
    StreetMAE & 78.49 & 81.97 & 71.54 & 76.40 \\
    StreetBYOL & 86.25 & 91.98 & 78.62 & 84.78 \\
    Seg+StreetBYOL & 87.42 & 91.03 & 82.27 & 86.43 \\
    \bottomrule
    \end{tabular}}
  \caption{Performance of different pre-train methods.}
  \label{table:pretrain}
\end{table}

% \begin{table}[h!]
%   \centering
%   \resizebox{1.0\linewidth}{!}{
%     \begin{tabular}{clcccc}
%     \toprule
%     \textbf{Freeze} & \textbf{Pre-training} & \textbf{Accuracy} & \textbf{Precision} & \textbf{Recall} & \textbf{F1-Score} \\
%     \midrule
%     \Checkmark & StreetMAE & 74.80  & 81.84  & 61.98  & 70.54 \\
%     \XSolidBrush & StreetMAE & 78.49 & 81.97 & 71.54 & 76.40 \\
%     \Checkmark & StreetBYOL & 83.48 & 91.02 & 73.28 & 81.19 \\
%     \XSolidBrush & StreetBYOL & 86.25 & 91.98 & 78.62 & 84.78 \\
%     \Checkmark & Seg+StreetBYOL & 84.48 & 91.35 & 75.25 & 82.52 \\
%     \XSolidBrush & Seg+StreetBYOL & 87.42 & 91.03 & 82.27 & 86.43 \\
%     \Checkmark & DINOv2 (ViT-B/14) & 88.20 & 92.08 & 82.88 & 87.24 \\
%     \XSolidBrush & DINOv2 (ViT-B/14) & \textbf{88.85}  & 92.77 & 83.62 & \textbf{87.96} \\
%     \bottomrule
%     \end{tabular}}
%   \caption{Best results of different pre-train methods.}
%   \label{table:pretrain}
% \end{table}
\begin{figure}[ht!]
    \centering
    \includegraphics[width=1.0\linewidth]{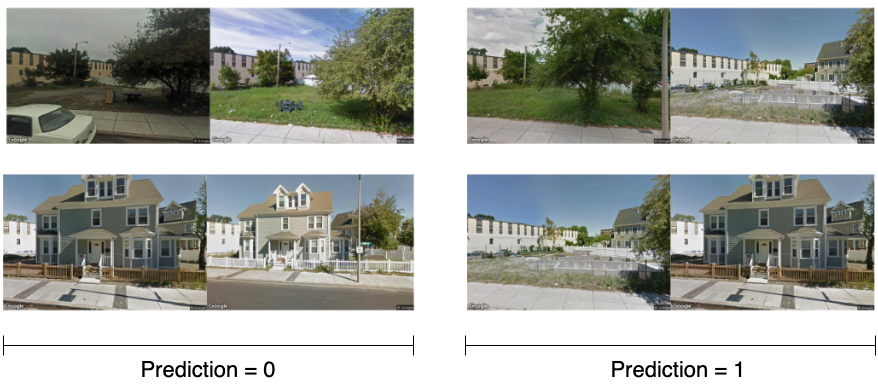}
    \caption{Sampled prediction results. Our proposed change detection model effectively identifies structural changes in buildings, while filtering our random variations such like lighting, shadows, vegetation, and vehicles.}
    \label{fig:res_noisy}
\end{figure}

% \begin{figure}[h!]
%     \centering
%     \includegraphics[width=1.0\linewidth]{figure/res_noise.png}
%     \caption{Sampled prediction results. Our proposed change detection model is able to filter out noisy signals such as changes in lighting, shades, vegetation, and vehicles and focus on change of building structures.}
%     \label{fig:res_noisy}
% \end{figure}
\begin{figure*}[h!]
    \centering
    \includegraphics[width=1.01\linewidth]{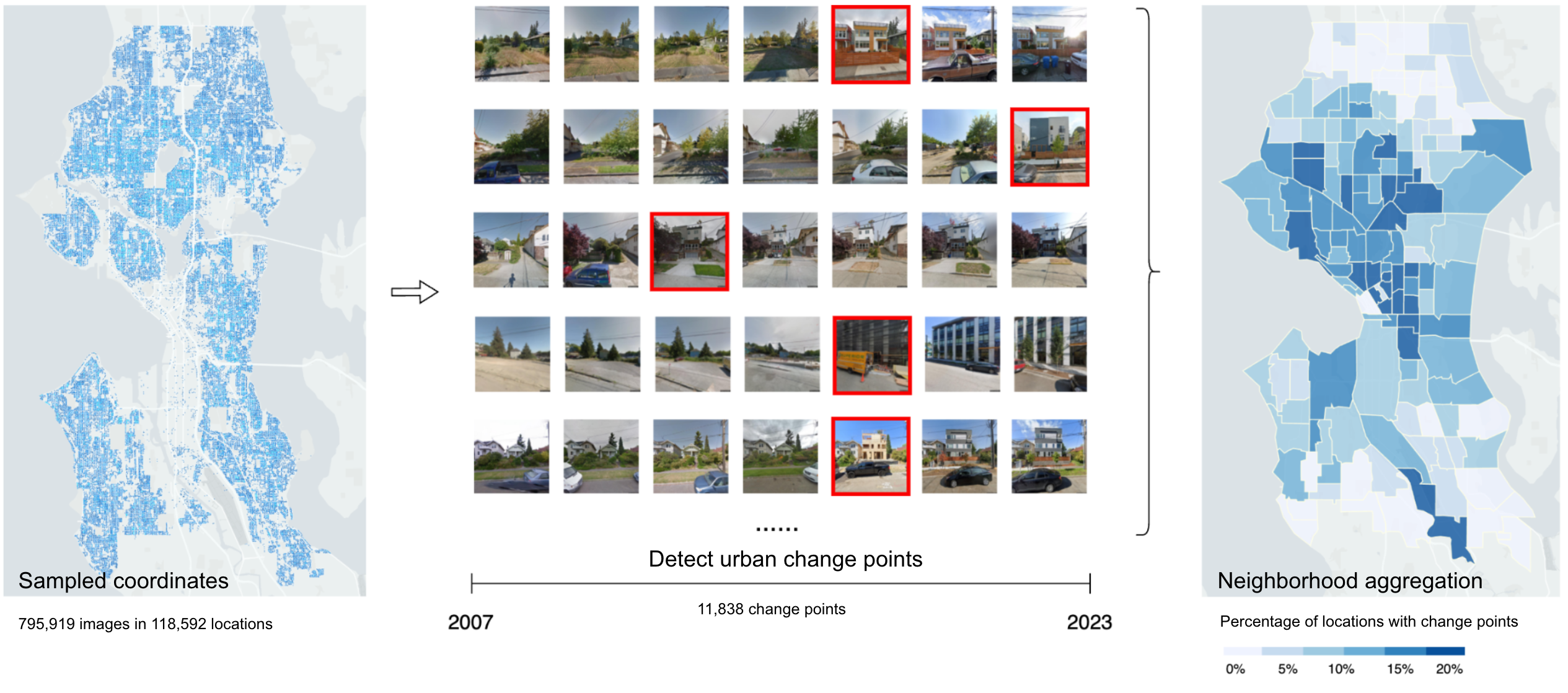}
    \caption{Assessing urban change in Seattle. \textbf{Left:} Location of approximately $800$k sampled street view images, each represented by a blue dot. \textbf{Middle:} Results from deploying our change detection model on the sampled images to pinpoint urban changes shown in red bounding boxes. \textbf{Right:} Change points, aggregated at the census tract level, with color denoting the proportion of street view time series that have been identified as change.}
    \label{fig:seattle}
\end{figure*}

% \begin{figure*}[h!]
%     \centering
%     \includegraphics[width=1.01\linewidth]{figure/seattle.png}
%     \caption{Assessing urban change in Seattle. \textbf{Left:} We sampled around $800k$ street view images in total and each blue dot indicates its location. \textbf{Middle:} Applying our proposed change detection model directly on all sampled images and identify urban change points. \textbf{Right:} Aggregating detected change points to the census tract level, the color indicates the percentage of sampled time series detected with urban change points.}
%     \label{fig:seattle}
% \end{figure*}
\section{Results and discussion}
As shown in Table \ref{table:backbones}, DINOv2 has demonstrated the best performance in our evaluation, achieving $88.85\%$ accuracy through fine-tuning. Notably, considering the presence of challenges like shadows and occlusions in the images, human performance in this change detection task is approximately around $90\%$ during our labeling process. This observation suggests that fine-tuning DINOv2 as the backbone network has enabled the model to approach human-level performance. Furthermore, freezing the DINOv2 network and training only the linear layers surpass the fine-tuning outcomes of all other backbone networks, strongly affirming the capacity of DINOv2 to generate potent visual features suitable for change detection. 
% It is worth noting that, when using patch features as the output and employing non-linear prediction heads, DINO achieves the best result as indicated in the table. On the other hand, DINOv2 shows distinct behavior; its optimal performance is attained when using the CLS Token as the output and employing a single linear layer as the prediction head.

We find the performance of the change detection model fine-tuned on the pairwise dataset is significantly lower than its performance attained after fine-tuning on our time series dataset, as illustrated in table \ref{table:timeseries}. 
Moreover, augmenting the dataset using artificial techniques can lead to adverse effects. 
% This phenomenon could be attributed to the introduction of additional noise and interference into the building elements of the street view images. Typically, artificial data augmentation involves transformations applied to the entire image, resulting in alterations in the appearance of buildings within the images, which subsequently disrupts the ability of the model to detect architectural changes.
As a form of natural data augmentation, time series data equips the model with sufficient information to identify and filter out irrelevant variations that occur over time, such as changes in lighting, vegetation, and vehicles as is shown in Figure \ref{fig:res_noisy}, which guides the model to focus on more temporally stable elements such as building structures. These results validate the critical role of our proposed time series data in the context of street view image change detection task.

% By progressively increasing the number of sampled image pairs from the sequences, we observe a gradual improvement in model performance. This underscores the essential significance of time series data. 
The performance of the street-view pre-trained models is presented in Table \ref{table:pretrain}. The results of StreetMAE are significantly lower compared to those of StreetBYOL. This may be because patch reconstruction process is more prone to learning color and texture information, which aligns with the noise we aim to eliminate in change detection, rather than building structure. The addition of the semantic segmentation module leads to a performance enhancement in StreetBYOL. Nevertheless, domain-specific pre-trained models, whether StreetMAE or StreetBYOL, do not surpass the performance of the generic visual model DINOv2. This can be attributed to the smaller training data size used for domain-specific pre-training compared to generic visual models. Specifically, the inherently noisy street view images, with their complex and cluttered scenes, make it difficult for models to grasp fundamental concepts like shape, location, and architecture from limited data.
% This could be attributed, on one hand, to the considerably smaller training data volume used for domain-specific pre-training compared to the vast data utilized by generic visual models. On the other hand, this is also caused by the inherently noisy street view images as training data, often comprising complex and cluttered scenes that make it challenging for models to learn fundamental concepts about shape, location, environment, and architecture from a limited set of data. 
% Given the relatively high annotation cost of street view images and the substantial amount of unannotated data that persists, the task of curating the data and designing larger-scale pre-training methods to enhance the accuracy of change detection remains a noteworthy avenue for exploration.

\section{Case study: Assessing urban change in Seattle}
% \textbf{Scope and promise for social impact: Likelihood of social impact is extremely high: the paper’s ideas are already being used in practice or could be immediately.}
To evaluate the generalizability of our proposed change detection model on a large scale, we prepare a large-scale street view time series data for the city of Seattle, Washington. Figure \ref{fig:seattle} demonstrates the sampling process. We then apply our change detection model to identify urban change points: In total, we detect $11,838$ change points from $795,919$ sampled images in Seattle. 
% Our data shows significant correlation with change in median household income and population size in each census tract.
% \begin{itemize}
%     \item visualization
%     sampled results (corner cases)
%     \item correlation benchmark with permits data
%     \item ACS prediction
% \end{itemize}
\subsubsection{Construction permits data.}
% We obtain permits data fro Seattle 
% As mentioned earlier, construction permits data are often adopted as a fine-grained proxy for urban change in literatures. To compare with our change detection model, We fetch construction permit data from the online permit center of the Seattle city government\footnote{Available at \url{https://data.seattle.gov/Permitting/Building-Permits}}. 
As previously noted, previous works frequently rely on construction permit data as a detailed proxy for urban evolution. To further compare them with the results from our proposed change detection model, we obtain construction permit data from the Seattle city government's online permit center.
Each entry in the permit data provides details such as the date of issuance, permit category, geospatial coordinates, estimated cost, and other requisite information as mandated by the government.
As a data pre-processing step, we keep the ``new'', ``alteration'' and ``addition'' categories to align with the definition of urban change points. Additionally, we curated a subset of permits that had a total estimated cost exceeding $\$100,000$ within a single year. This approach allows us to compare our findings with both the complete permit dataset and the high-value permits, the latter of which are more probable to signify visible physical alterations.

% This way, urban change recorded by our filtered permits is significant enough to be viewed as potential signals of physical change.
\begin{figure*}[h!]
    \centering
    \includegraphics[width=0.93\linewidth]{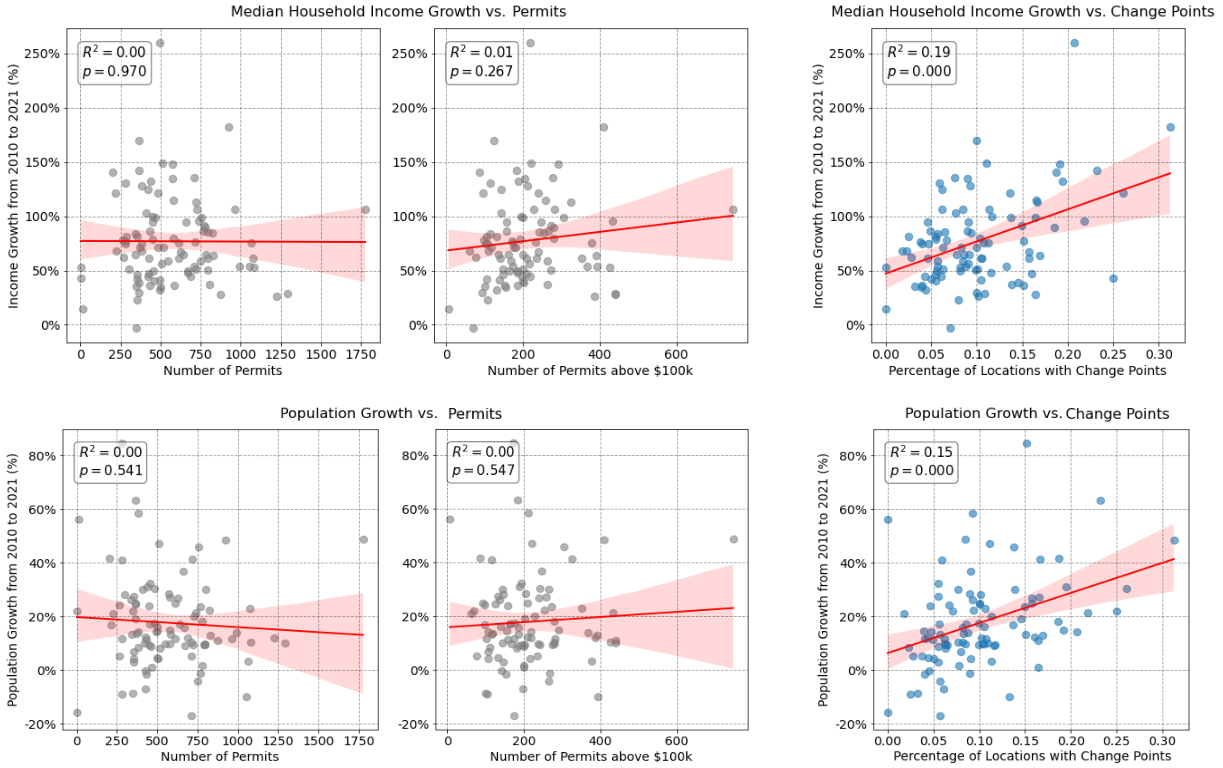}
    \caption{Linear correlation with socio-demographic indicators. \textbf{Top:} Median household income. \textbf{Bottom:} Population size. Each dot represents a Seattle census tract. The change detection results show statistically significant correlations with socio-demographic metrics, in contrast to construction permit data which lacks such correlation.}
    \label{fig:acs}
\end{figure*}

% \begin{figure*}[h!]
%     \centering
%     \includegraphics[width=0.9\linewidth]{figure/acs.png}
%     \caption{Correlation of change detection results with socio-demographic indicators. \textbf{Top:} Median household income. \textbf{Bottom:} Population size. Each dot corresponds to a Seattle census tract. Notably, these results exhibit significant correlations with socio-demographic metrics, unlike the construction permit data.}
%     \label{fig:acs}
% \end{figure*}

% \begin{figure*}[h!]
%     \centering
%     \includegraphics[width=0.96\linewidth]{./figure/acs.png}
%     \caption{Linear correlation with social-demographic data. \textbf{Top:} Median household income. \textbf{Bottom: } Population size. Each dot represents a census tract in Seattle. Change detection results demonstrates statiscally signifcant correlations with social-demographic data, while construction permit data fail to do so.
%     }
%     % between ACS attributes and total number of predicted change points in each census tract vs. correlation between ACS attributes and total number of construction permits in each census tract}
%     \label{fig:acs}
% \end{figure*}
\subsubsection{Correlation with social-demographic data.} 
We prepare social-demographic data at the census tract level from the American Community Survey (ACS) 5-year estimates. Specifically, we select population size and median household incomes as our target variables, and calculate their relative percentage change from 2009 to 2021 for each census tract in Seattle. 
To quantify the linear correlation between proxies for urban change and shifts in socio-demographics, we compare three distinct proxies: the entire set of permits, high-value permits (those exceeding $\$100$k), and the percentage of locations with urban change points identified using our proposed methodology.
As is shown in Figure \ref{fig:acs}, both the entire set of permits and the high-value permits fail to show a linear correlation with the change of median household income and population size in each census tract with $p$ value larger than $0.05$ and $R^2$ close to $0$. 
While the change points results from our proposed method reach an $R^2$ of $0.19$ and $0.15$ for median household income change and population size change respectively, and achieve a $p$ value much less than $0.05$ supporting the statistical significance.
These results not only indicate that the detected urban change points provide a more accurate assessment of real-world urban transformations and socio-economic shifts, but also validates that our proposed change detection model can effectively complement existing proxies as a credible indicator of urban change.

\section{Conclusion and Future Work}
In this work, we propose a framework to assess fine-grained urban change at scale with street view time series. We have curated the largest street-level scene change detection dataset by far, and proposed an end-to-end change detection pipeline to identify urban change points at scale. We validate the proposed model by correlating with social-demographic data and prove its potential as a high-definition, up-to-date, and on-the-ground visual proxy of urban change.

While our data-driven approach provides a novel method to assess urban change, it is still subjective to a few limitations: 1) Street view data focus on changes observable at the street level, excluding alterations that might be non-visible, such as interior renovations. 2) The spatial-temporal distribution of Google Street View data is not consistent. Since its debut in 2007, Google has frequently updated its imagery in countries such as the US, but has been less consistent in updating images in many other regions, especially in developing countries. 
Despite these limitations, we believe our proposed method offer a comprehensive and extensive resource for urban change detection task, helping expand its social impact and advance sustainable development goals. 
% More broadly, identifying fine-grained urban change will benefit urban planing and advancing sustainable development goals. 
In future works, we can explore multi-task models to identify changes in a wider array of objects, 
% Additionally, our method can be integrated with satellite change detection systems, 
enhancing the applicability to a broader range of downstream tasks in cities.

\section{Acknowledgments}
This project was supported by the Google Cloud Grant from the Stanford Institute for Human-Centered Artificial Intelligence. The author would like to thank Zhecheng Wang, Sarthak Kanodia and Timothy Dai for their extensive guidance.
\bibliography{aaai24}

\end{document}